\pdfoutput=1
    \pdfoutput=1
    \pdfoutput=1
    \pdfoutput=1
    \pdfoutput=1
    \def\year{2020}\relax
    \documentclass[letterpaper]{article} 
    \usepackage{aaai20}  
    \usepackage{times}  
    \usepackage{helvet} 
    \usepackage{courier}  
    \usepackage{graphicx} 
    \usepackage{epsfig}
    \usepackage{amsmath}
    \usepackage{amssymb}
    \usepackage{float}
    \usepackage{multirow}
    \usepackage{multicol}
    \usepackage{verbatim}
    \usepackage{natbib}
    \title{Structure-Attentioned Memory Network for Monocular Depth Estimation }
   \author{Jing Zhu$^{1,2,3}$ \hspace{0.5cm} Yunxiao Shi$^{1,2}$ \hspace{0.5cm} Mengwei Ren$^{1,2}$ \hspace{0.5cm} Yi Fang$^{1,2,3}$\thanks{indicates corresponding author} \hspace{0.5cm} Kuo-Chin Lien$^{4}$ \hspace{0.5cm} Junli Gu$^{4}$  \vspace{0.2cm}\\
   $^{1}$NYU Multimedia and Visual Computing Lab, USA \\
$^{2}$New York University, USA \\
$^{3}$New York University Abu Dhabi, UAE\\
$^{4}$XMotors.ai \\
{\tt\small {\{jingzhu, yunxiao.shi, mengwei.ren, yfang\}@nyu.edu} \hspace{0.5cm} \{kuochin, junli\}@xmotors.ai}} 
    \usepackage{blindtext}

	\usepackage{etoolbox}
	\makeatletter
	\patchcmd{\maketitle}{\@copyrightspace}{}{}{}
	\makeatother
     
\begin{document}
    
    \maketitle
    
    \begin{abstract}
 Monocular depth estimation is a challenging task that aims to predict a corresponding depth map from a given single RGB image. Recent deep learning models have been proposed to predict the depth from the image by learning the alignment of deep features between the RGB image and the depth domains. In this paper, we present a novel approach, named Structure-Attentioned Memory Network, to more effectively transfer domain features for monocular depth estimation by taking into account the common structure regularities (e.g., repetitive  structure  patterns,  planar  surfaces, symmetries) in domain adaptation. To this end, we introduce a new Structure-Oriented Memory (SOM) module to learn and memorize the structure-specific information between RGB image domain and the depth domain. More specifically, in the SOM module, we develop a Memorable Bank of Filters (MBF) unit to learn a set of filters that memorize the structure-aware image-depth residual pattern, and also an Attention Guided Controller (AGC) unit to control the filter selection in the MBF given image features queries. Given the query image feature, the trained SOM module is able to adaptively select the best customized filters for cross-domain feature transferring with an optimal structural disparity between image and depth. In summary, we focus on addressing this structure-specific domain adaption challenge by proposing a novel end-to-end multi-scale memorable network for monocular depth estimation. The experiments show that our proposed model demonstrates the superior performance compared to the existing supervised monocular depth estimation approaches on the challenging KITTI and NYU Depth V2 benchmarks.
\end{abstract}
    
    \section{Introduction} 
    \begin{figure}[t]
    \begin{center}
       \includegraphics[width=\linewidth,height = 3.6cm]{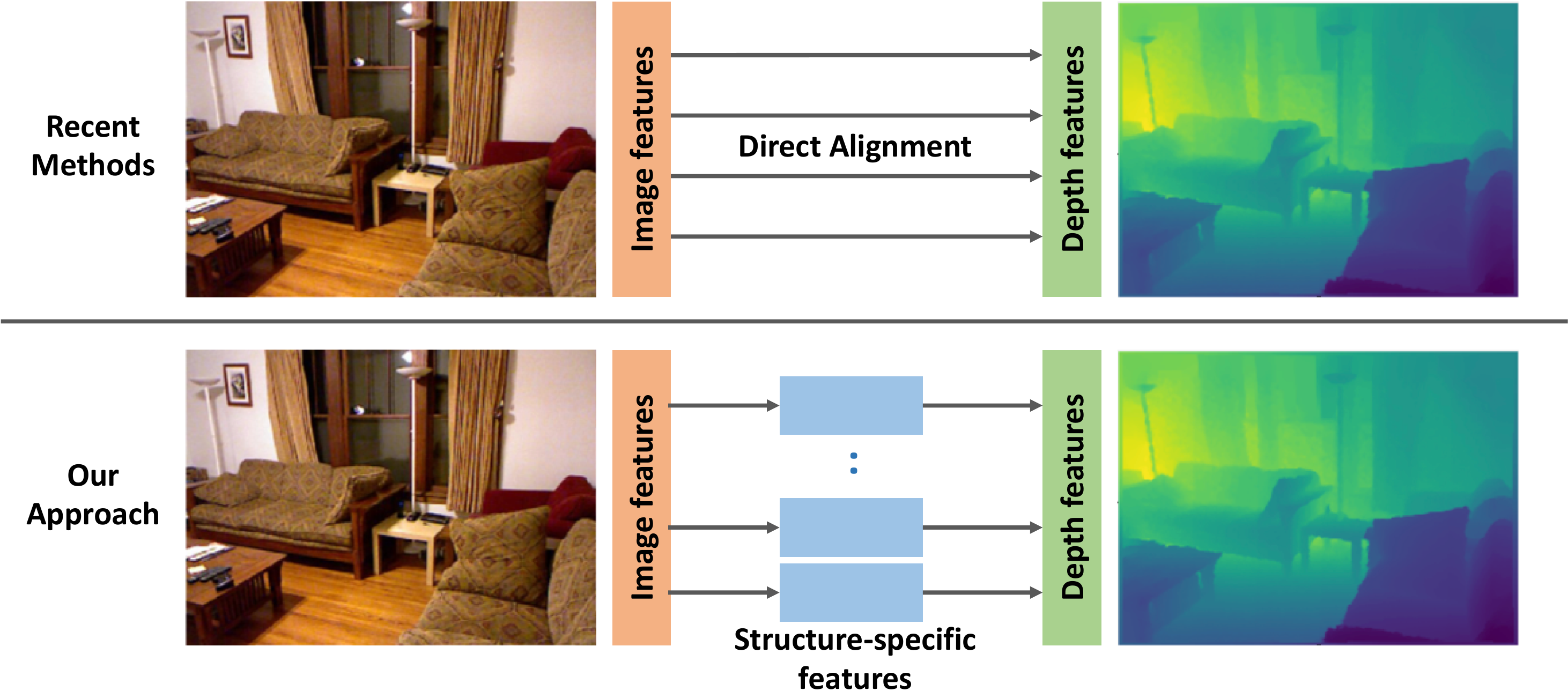}
    \end{center}
    \vspace{-11pt}
       \caption{Different from the recent methods that directly align the features from different domains, we focus on the structure-specific domain adaption.}
       \vspace{-12pt}
    \label{DiffFig}
    \end{figure}
    Depth estimation is an important component in many 3D computer vision tasks like visual Simultaneous Localization and Mapping (visual SLAM). Traditional approaches have made significant progress in binocular or multi-view depth estimation by taking advantage of geometry constraints of either spatial (i.e. stereo camera) or temporal (i.e. video sequence) pairs. With the prevalence of deep convolutional neural networks, researchers have been trying to relax the constraints by tackling monocular depth estimation. Recent works (\cite{Wang2015Towards,Roy2016Monocular,Kuznietsov2017Semi,Kim2016Unified,Fu2018Deep,Eigen2015Predicting}) have demonstrated promising results using regression-based deep learning models. Their models are trained by minimizing image-level losses with supervised signal on predicted results. Nevertheless, the cross-modality variance between the RGB image and the depth map still makes monocular depth prediction an ill-posed problem. Based on this observation, some researchers have considered solving the problem with additional feature-level structural constraints by minimizing the cross-modality residual complexity between image features and depth features. Most existing methods either consider the pixel-wise or structure-wise alignment in this regard. For instance, several architectures utilize the micro discrepancy loss as similarity measures such like sum of squared differences, correlation coefficients (\cite{Myronenko2010Intensity}) and maximum mean discrepancy (\cite{Ghifary2015Domain,Long2015Learning}) to align the RGB images features with depth features from pixel to pixel independently without considering the spatial dependencies. Another line of work has tried to apply the adversarial adaptation methods (\cite{Kundu2018AdaDepth,Tzeng2017Adversarial,Judy2015Simultaneous}) in conjunction with task-specific losses that concentrate on macro spatial distribution similarity between the image features and depth ones. In this paper, we seek a way to address this domain adaption challenge on both pixel-wise discrepancies and the structure dependencies by extracting the structure-specific information between the two domains (as shown in Figure \ref{DiffFig}).\\
        \indent In order to explore the pixel-wise discrepancies as well as the structure dependencies between the image features and depth features, we propose a memorable domain adaptation network, with an image-encoder-depth-decoder regression network backbone, and a specifically designed Structure-Oriented Memory (SOM) module coupled with a cross-modality residual complexity loss to minimize the gap between latent distribution of the image and depth map from both the pixel-level and structure-level. Given the observation that similar type of scenes (e.g. roadside scenes) often share common structural regularities (e.g. repetitive structure patterns, planar surfaces, symmetries), a set of filters could be trained to learn a specific structural image-depth residual patterns. Therefore, in our SOM module, we build a Memorable Bank of Filters (MBF) to store and learn the structure-ware filters, then we construct an Attention Guided Controller (AGC) to learn to automatically select the appropriate filters (from the MBF) to capture the significant information from the given image features (generated by the image encoder) for the further depth estimation. Finally, the customized image features are fed into the depth decoder network to output the corresponding depth maps. Importantly, comparing to the direct alignment between the two domains features (e.g. direct applying $L_1$ loss between $Z_i$ and $Z_d$), our introduced SOM module not only improves the fitting ability, but also reduces the training burden of the image encoder simultaneously. The experiments conducted on two well-known large scale benchmarks KITTI and NYU Depth V2, demonstrate that our proposed model obtains the state-of-the-art performance on monocular depth estimation tasks. Moreover, the performance margin between model trained with SOM and the one trained with direct alignment, validate the effectiveness of our proposed SOM module.
        \begin{figure*}
    \begin{center}
    \includegraphics[width=0.85\textwidth, height = 6cm]{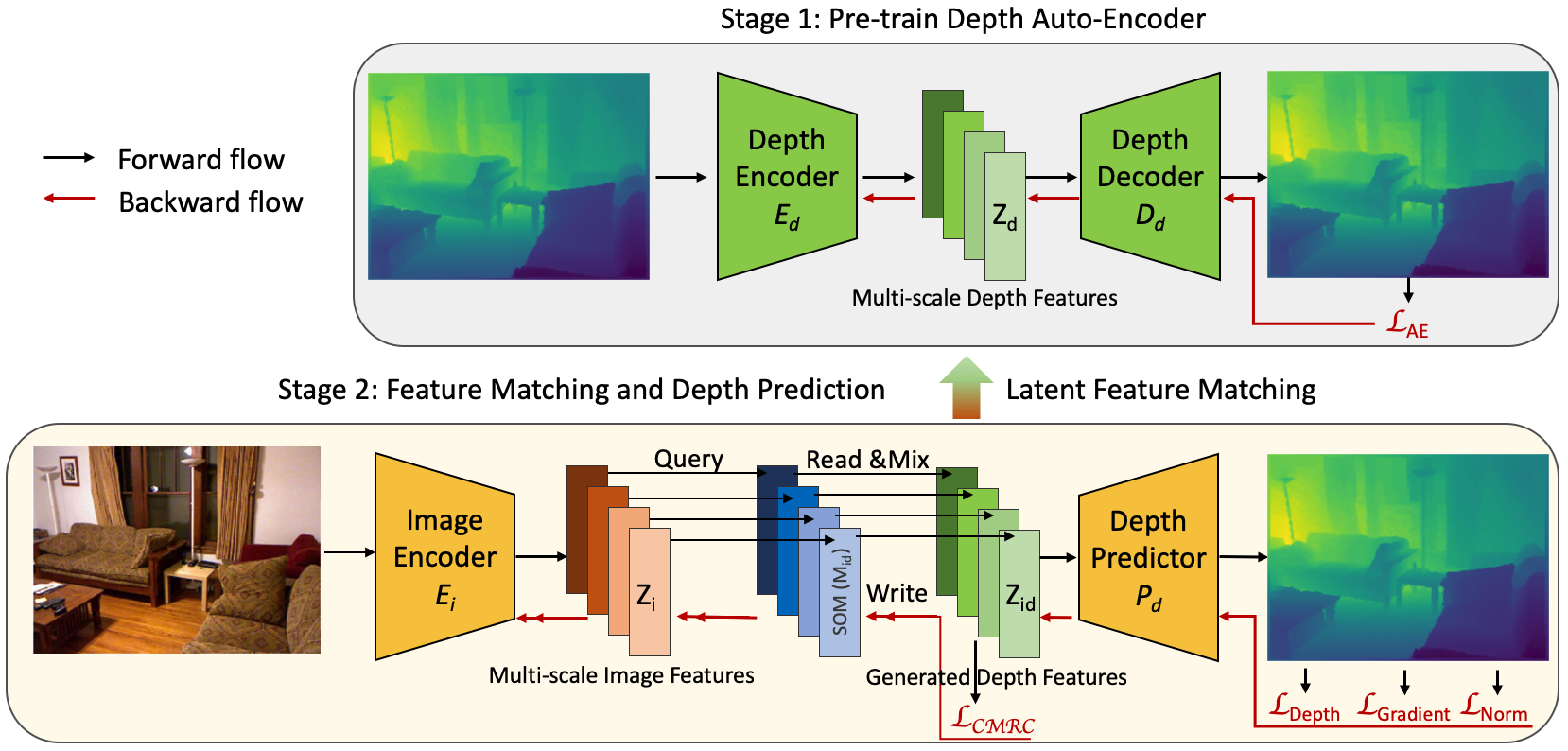}
    \end{center}
    \vspace{-15pt}
       \caption{The pipeline of our proposed Structure-Attentioned Memory Network. }
    \label{fig:pipeline}
    \vspace{-10pt}
    \end{figure*}
    In summary, our contributions in this paper are as follows:
    \begin{itemize}
    \item We introduce memory strategies to address monocular depth estimation by designing a novel Structure-Oriented Memory (SOM) module with a Memorable Bank of Filters (MBF) and an Attention Guided Controller (AGC) for feature-level cross-modality domain adaptation.
      \item We propose a novel end-to-end deep learning Structure-Attentioned Memory Network, which seamlessly integrates a front-end regression network with the SOM module that operates at feature-level to substantially improve the depth prediction performance. 
    \item We achieve state-of-the-art performance on two large scale benchmarks: KITTI and NYU Depth V2, which validates the effectiveness of the proposed method. 
    \end{itemize}
    
    The remainder of our paper is organized as follows. We present a brief review of the related literature in Section \emph{Related Works}, after which we introduce the proposed method in details in Section \emph{Proposed Method}. In Section \emph{Experiments}, we provide the qualitative and quantitative experimental results, as well as ablation studies that demonstrate the effectiveness of the proposed method. Finally, we conclude the paper in Section \emph{Conclusion}. 
    
    \label{sec:related}
    \section{Related Works}
     Monocular depth estimation is a fundamental problem in computer vision which has widespread application in graphics, robotics and AR/VR. While previous works mainly tackle this using hand-crafted image features or probabilistic models such as Markov Random Fields (MRFs) (\cite{Saxena2009Make3D}), recent success of deep learning based methods (\cite{Wang2015Towards, Roy2016Monocular,Kuznietsov2017Semi,Kim2016Unified,Fu2018Deep, Eigen2015Predicting}) have inspired researchers to use deep learning techniques to address the challenging depth estimation problem. The learning based monocular depth estimation approaches can be mainly summarized into two categories, the supervised and the unsupervised/semi-supervised methods.
    
    \textbf{Supervised Methods} \hspace{0.3cm} A majority of works focus on supervised learning to use the learned features from CNNs to do accurate depth prediction. \cite{Eigen2014Depth} first brought CNNs to depth regression task by integrating coarse and refined features with a two-stage network. The multi-task learning strategies were also applied in depth estimation to boost the performance. \cite{Liu2010Single} utilized the semantic segmentation as objectness cues for depth estimation. Furthermore, \cite{Shi2014Pulling} and \cite{Xu2018PAD} performed joint prediction of the pixel-level semantic labels as well as the depth. Surface normal information was also adopted in many recent works (\cite{Eigen2015Predicting,Zhou2017Unsupervised,Wang2015Towards,Qi2018GeoNetG}). Besides, some research works also demonstrated the robustness of multi-scale feature fusion in pixel-level prediction tasks (e.g. semantic segmentation, depth estimation). \cite{Fu2018Deep} adopted the dilated convolution to enlarge the perceptive field without decreasing spatial resolution of the feature maps. In \cite{Buyssens2012Multiscale}'s work, inputs at different resolutions are utilized to build a multi-stream architecture. Instead of regression, there are also methods that discretize the depth range and transfer the regression problem to a classification problem. In the work of \cite{Fu2018Deep},  the space-increasing discretization is proposed to reduce the over-strengthened loss for the large depth values.
    
    \textbf{Unsupervised/Semi-supervised Methods} \hspace{0.3cm}  Another line of methods on monocular image depth prediction goes along the unsupervised/semi-supervised direction which mostly takes advantage of geometry constraints (e.g. epipolar geometry) on either spatial (between left-right pairs) or temporal (forward-backward) relationship. \cite{Garg2016Unsupervised} proposed to estimate the depth map from a pair of stereo images by imposing the left-right consistency loss. \cite{Zhan2018Unsupervised} jointly learned a single view depth estimator and monocular odometry estimator using stereo video sequences, which enables the use of both spatial and temporal photometric warp constraints. Moreover, following the trend of adversarial learning, the generative adversarial networks (GANs) have been utilized in the depth estimation problem. \cite{Kundu2018AdaDepth} proposed an unsupervised domain adaptation strategy for adapting depth predictions from synthetic RGB-D pairs to natural scenes in the depth estimation task.
    
    \textbf{Cross-Modality Domain Adaptation} \hspace{0.3cm} In addition to the recent depth estimation methods, research works focused on the cross-modality domain adaption are also highly relevant to ours. The existence of cross modality, or domain shift, is commonly seen in real-world application, which is the consequence of data captured by different sensors (e.g. optical camera, LiDAR or stereo camera), or varying conditions (i.e. background). In different domains, semantic labels are shared whereas the data distributions are usually different to a large extent. For example, Computed Tomography (CT) and Magnetic Resonance Imaging (MRI) in bio-medical image analysis (\cite{Dou2018Unsupervised}); RGB images, depth maps and point clouds in 2D, 2.5D and 3D computer vision tasks. Numerous approaches have been proposed to address the domain adaptation needs in different visual tasks. Here we briefly review some domain adaption methods using deep learning techniques.
    
    Most deep domain adaptation methods utilize a siamese architecture with two streams for source and target models respectively, and the network is trained with a discrepancy loss to minimize the pixel-wise shift between domains. \cite{Long2015Learning} used maximum mean discrepancy together with a task-specific loss to adapt the source and target, while \cite{Sun2016Deep} proposed the deep correlation alignment algorithm to match the mean and covariance. \cite{Bloesch_2018_CVPR} proposed to learn a dense representation using an auto-encoder. \cite{Mandikal20183D} trained the network with $L_1$ constrain in latent space to transfer feature from 2D to 3D in order to directly predict 3D point cloud from a single image. In our work, we aim to design a domain adaptive (SOM) module using memory mechanism, so that the image features can be automatically customized to obtain a better depth prediction.

    \section{Proposed Method}
    \label{sec:proposed}
     The monocular depth estimation problem can be defined as a nonlinear mapping $f: I \rightarrow Y$ from the RGB image $I$ to the geometric depth map $Y$, which can be learned in a supervised fashion given a training set $X = \{I^t,Y^t\}_{t=1}^N$. To learn the mapping function, we propose Structure-Attentioned Memory Network as shown in Figure \ref{fig:pipeline}, which is composed of a (pre-trained) depth auto-encoder, an image encoder and a depth predictor equipped with SOM module. All the components are trained into two stages. In the first stage, a series of `target' depth features $\{Z_d^t\}_{t=1}^k \in R^k$ are learned by training a depth map auto-encoder $(E_d, D_d)$. In the second stage, we train an image encoder $E_i$, SOM modules $M_{id}$ and a depth predictor $P_d$ to map the 2D image to the depth map in an end-to-end manner. Particularly, $E_i$ encodes the RGB image to the `source' image features $\{Z_i^t\}_{t=1}^k \in R^k$, which act as queries to obtain image-depth residual patterns from SOM module. The residual is then concatenated to the source feature to form a newly transferred feature set $\{Z_{id}^t\}_{t=1}^k \in R^k$ (which is expected to be aligned with the target feature $\{Z_d^t\}_{t=1}^k$ with supervision) is fed to the predictor $P_d$ to estimate the output depth map. We will elaborate the network structures  from two stages separately. 
    \begin{figure*}[t]
    \begin{center}
       \includegraphics[width=0.9\linewidth, height = 6.3cm]{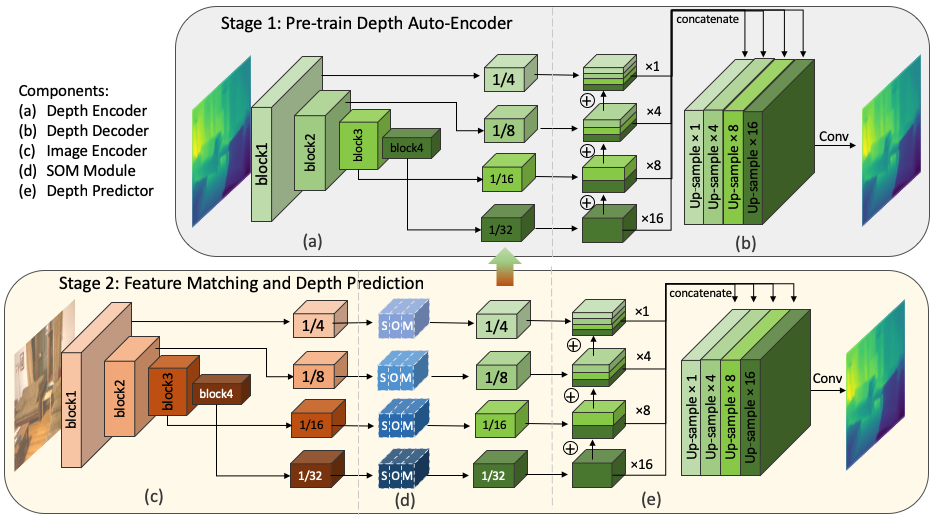}
    \end{center}
    \vspace{-13pt}
       \caption{The network structure of Structure-Attentioned Memory Network.}
       \vspace{-10pt}
    \label{fig:network}
    \end{figure*}
    
    \subsection{Stage 1: Depth Auto-Encoder}
    \label{subsec:ae}
    In order to learn a strong and robust prior over the depth map as a reference in the latent matching process, we train a depth auto-encoder $(E_d, D_d)$ which takes a ground truth depth map $Y_d\in R^{M \times N}$ as input, and outputs a reconstructed depth map $\hat{Y}_d\in R^{M \times N}$. As shown in Figure \ref{fig:network} (Stage 1), we use the DenseNet based encoder-decoder structure. Specifically, DenseNet-121 is utilized for constructing the depth encoder (Figure \ref{fig:network} (a)), in which four feature maps with cascading resolutions are extracted from different blocks (shallow to deep) for depth decoding. In order to make sure that the object contours as well as details are well preserved, we use a Feature Pyramid Network (FPN) to build the depth decoder, fusing multi-scale features in a pyramid structure. Specifically, as shown in Figure \ref{fig:network} (b), four features with sizes $1/4$, $1/8$, $1/16$ and $1/32$ of the input are derived. Starting from the deepest feature, each feature map is first upsampled by a factor of $2$, and element-wisely added to its following feature map. After the fusion process of the multi-scale feature maps, each of the newly generated feature maps is upsampled to size of $1/4$ the original input (or the size of the shallowest feature map), and concatenated together to form a feature volume. Finally, the output depth map is predicted via extra CNN layers on the concatenated feature volume. The FPN decoder is able to preserve details in the depth map decoding process. We will show more experimental comparison between different decoder structure to demonstrate its effectiveness in the \emph{Experiments} section.
    
    \subsection{Stage 2: Depth Prediction with SOM Module for Latent Space Adaptation}
     In the second stage, we aim to train the network in an end-to-end manner to effectively transfer the features derived from image encoder $E_i$ from image domain to depth domain, as a strong prior over the ground truth depth, so as to better deduce the depth from the transferred prior. To this end, this stage contains three major components as shown in Figure \ref{fig:network} (c), (d) and (e): the image encoder, the SOM module for latent space adaptation, and the depth predictor $(E_i, M_{id}, P_d)$. Each component of the network will be explained below. 
    
    \textbf{Image Encoder and Depth Predictor as Regression Backbone} \hspace{0.3cm} In order to make sure that the network derive both depth features and image features at the same scale, we design the encoder-decoder based backbone ((c) and (d) in Figure \ref{fig:network}) for stage 2 exactly the same as those of stage 1 but without weight sharing. Specifically, the structure of image encoder $E_i$ ((c) in Figure \ref{fig:network}) is identical to that of depth encoder $E_d$ ((e) in Figure \ref{fig:network}), and similarly for $D_d$ ((b)) and $P_d$ ((e)). 
    
\begin{figure}[t]
    	\centering
    	\includegraphics[width=0.9\linewidth, height = 6.5cm]{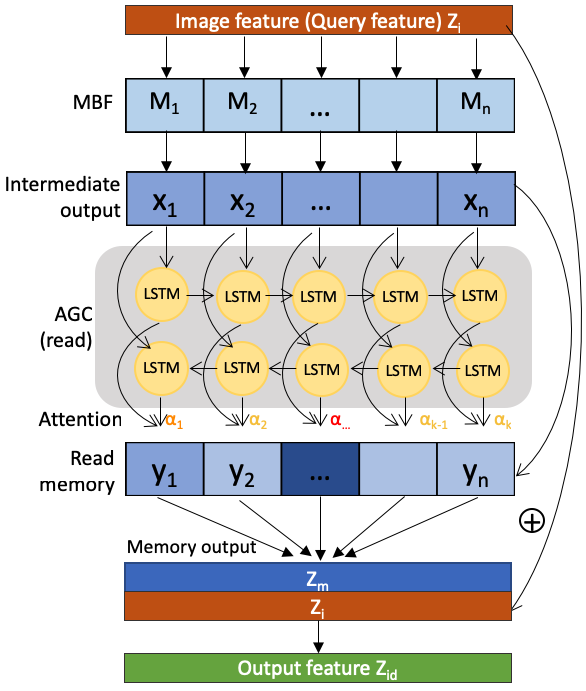}
    	\vspace{-10pt}
    	\caption{The SOM reading process (of a single SOM module).}
    	\label{fig:memory}
    	\vspace{-15pt}
    \end{figure}

    \textbf{SOM Module for Latent Space Adaptation} \hspace{0.3cm} In the latent space, we propose an additional structure oriented memory module consisting of two collaborative units: a Memorable Bank of Filters (MBF) that stores a bank of learned filters to detect the cross-modality residual complexity between the depth feature and the image feature, and an Attention Guided Controller (AGC) which controls the interaction between the image feature with the MBF. The image feature as a specific query feature selects filters from MBF with an attention guided read controller, and the MBF is updated through a write controller that is naturally integrated into the back propagation to make the network can be trained end-to-end. The proposed SOM reading and writing process are as follows.\\
    \indent\textbf{\textit{SOM Reading}} Different from reading by `addressing' in general memory concept, the proposed SOM module is reading by `attention', which means each memory slot is assigned with a weight, and the whole memory is merged per weights as reading output. As demonstrated in Figure \ref{fig:memory}, given the query feature $Z_i$, in order to obtain weights for each memory slot, we build a LSTM-based read controller to learn the weights. Specifically, each filter from the memory slot $\{M_t\}_{t=1}^n$ is firstly convolved on the feature, and the intermediate outputs are denoted as  $\{x_t\}_{t=1}^n$, where $n$ is the memory size, and $x_t$ is formulated as:
    $x_t = W_t * Z_i + b_t, M_t = (W_t,b_t)$, $W_t$ is the kernel, $b_t$ is the bias, and $*$ is the convolution operation. The intermediate outputs $\{x_t\}_{t=1}^n$ could be thought of as the `unweighted/unbiased' output that takes each filter/memory slot equally. Then in order to further add weighted attention on the result pool, a Bi-Directional Convolutional Long Short Term Memory is applied as the read controller on $\{x_t\}_{t=1}^n$ to explore the correlation within the pool, so as to aggregate the memory slots with strong attention. Particularly, read controller processes $\{x_t\}_{t=1}^n$ from two directions and computes the forward hidden sequence $h_f$ by iterating the input from $t=1$ to $n$, and the backward hidden sequence $h_b$ by iterating the input from $t=n$ to $1$. The forward/backward flow of the LSTM cell is formulated as below: 
    $\begin{array}{ll}
    i_t = \sigma(W_{xi}\ast x_t + W_{hi}\ast h_{t-1} + W_{ci}\circ c_{t-1} + b_i)\\
    f_t = \sigma(W_{xf}\ast x_t + W_{hf}\ast h_{t-1} + W_{cf}\circ c_{t-1} + b_f)\\
    c_t = f_t \circ c_{t-1} + i_t \circ \tanh(W_{xc} \ast x_t + W_{hc} * h_{t-1} + b_c)\\
    o_t = \sigma(W_{xo} * x_t + W_{ho}\ast h_{t-1} + W_{co} \circ c_t + b_o)\\
    h_t = o_t \circ \tanh(c_t)
    \end{array}$
    where $h$ is the hidden sequence, $\sigma$ is the logistic sigmoid function, $\ast$ is the convolution operator and $\circ$ denotes the Hadamard product. $i_t, f_t, o_t, c_t$ represent input gate, forget gate, output gate, and cell activation vector respectively,  and $W_{hi}$ is the hidden-input gate matrix, while $W_{xo}$ is the input-output gate matrix.
    The final attention sequence $\alpha$ is computed with regard to both $h_f$ and $h_b$ as follows:
    $\alpha_t = softmax(W_{{h_f y}}{h_{f(t)}} + W_{{h_b}y}{h_{b(t)}} + b_y ) $,
    where $t=1$ to $n$, and each $y$ after softmax operation in the output sequence is associated with the weight for each memory slot  (refer to $\alpha$ value in Figure \ref{fig:memory}, the redder the color, the higher the attention), therefore $\sum_{i=1}^k\alpha_i=1$. The memory output $Z_m$ is a combination of the output sequence that focuses more on the slot with higher attention, while less on lower attention value:  $Z_m = \sum_{t=1}^n y_t, y_t = \alpha_t x_t $.
    Finally, $Z_m$ is concatenated with the query feature itself to reproduce a transferred feature $Z_{id}$ that is supposed to match the distribution of the depth feature $Z_d$. \\
    \indent\textbf{\textit{SOM Writing}} The proposed memory writer can be seamlessly integrated to network back propagation. The attention learned from the read controller will also operate in the memory writing process, and specifically, the slot with higher attention will be updated to a larger extent and vice versa. As shown in Figure \ref{fig:pipeline}, there are two backward flows that affect the writing of the memory (red arrows in Figure \ref{fig:pipeline}): one comes from the output branch, and the other comes from the latent matching branch. The update rule could be formulated (in a simplified form) as 
    $W_t \leftarrow W_t + \alpha_t \eta\Delta_{W_t}$, where $\alpha_t$ is the attention for each slot, $\eta$ is the learning rate, and $\Delta_{W_t}$ is the total gradient from both branches.
    
    \subsection{Learning objectives}
    We design multiple objectives to constrain the joint training of the network with details as follows.
    
    \textbf{Depth Estimation Objective} \hspace{0.3cm} The depth estimation objective poses constraints on the front-end pipeline of the single image depth estimation. A common way for supervising regression tasks is to adopt $L_1$ or $L_2$ loss between the prediction and the ground truth, which means that larger values have much heavier influence on the loss. However, in depth estimation task, the larger the depth value is, the farther the object is to the camera, which means that the information is less rich for the estimator, leading to unnecessarily large loss (\cite{Fu2018Deep}). Therefore, in order to reduce the over-emphasized error on large depth values, we use the logarithm mean squared error ($\text{RMSE}_{\log}$)) loss to make the predictor focus more on closer objects which makes up the main portion in a depth map. The objective is formulated as $\mathcal{L}_{depth} = \sqrt{\frac{1}{N}\sum_{i\in N}||\log(d_i) - \log(d_i^*)||^2}$, where $d$ is the ground truth depth map, while $d^*$ is the predicted depth map.
    
    \textbf{Auto-Encoder Objective} \hspace{0.3cm} The objective for the depth auto-encoder is utilized in the first training stage. To make sure that the depth features and the image features are in the same scale with same constraints, we also applied the $\text{RMSE}_{\log}$ on the auto-encoder as $\mathcal{L}_{AE} = \sqrt{\frac{1}{N}\sum_{i\in N}||\log(d_i) - \log(\hat d_i)||^2}$, where $d$ is the ground truth depth map, while $\hat d$ is the reconstructed depth map. 
    
    \textbf{Cross-Modality Residual Complexity Objective} \hspace{0.3cm} The latent adaptation objective is applied to constrain the SOM module to minimize feature distribution discrepancies. We use $L_1$ loss between the `target' depth features (pretrained from stage 1) and the SOM transferred image features. The objective is a sum of feature alignment losses at different levels as $\mathcal{L}_{CMRC} = \sum_k||Z^k_{id} - Z^k_d||_1$, where $k$ is the number of features involved in latent matching. 
    
    \textbf{Gradient and Surface Normal Constraints} \hspace{0.3cm} To further strengthen the network by pulling out the model from local minima, we added extra constraints on the predicted depth map including the gradient loss and the surface normal loss to finetune the training following commonly used techniques. The gradient loss is defined as $\mathcal{L}_{gradient} = \frac{1}{N}\sum_{i=1}^N ||\nabla d_i - \nabla d_i^* ||_1$, and specifically, we adopt Sobel filter to calculate the gradient both vertically and horizontally; $\nabla d$ is the image gradient of the ground truth depth map, while $\nabla d^*$ is the image gradient of the predicted depth map. 
    The surface normal loss is defined as the similarity between the surface normal of the ground truth depth map with the predicted depth map as $\mathcal{L}_{normal} = \frac{1}{N} \sum_{i=1}^N {( 1 - {\frac{<\nabla d_i, \nabla d_i^*>}{||\nabla d_i||_2 ||\nabla d_i^*||_2}})}$, formulated with the corresponding gradient. 
    
    In total, the training objectives are summarized as follows: (1) In training stage 1, the total loss is:
    $\mathcal{L}_{S_1} = \mathcal{L}_{AE}$; (2) In training stage 2, the total loss is a weighted sum of $\mathcal{L}_{depth}$, $\mathcal{L}_{CMRC}$, $\mathcal{L}_{gradient}$ and $\mathcal{L}_{normal}$, which is formulated as: $\mathcal{L}_{S_2} = \lambda_{depth}\mathcal{L}_{depth} + \lambda_{CMRC}\mathcal{L}_{CMRC} + \lambda_{gradient}\mathcal{L}_{gradient} + \lambda_{normal} \mathcal{L}_{normal}$, where $\lambda$ is the weight for each objective.
    
    \begin{figure*}[t]
    \begin{center}
    \includegraphics[width=\textwidth, height=4.85cm]{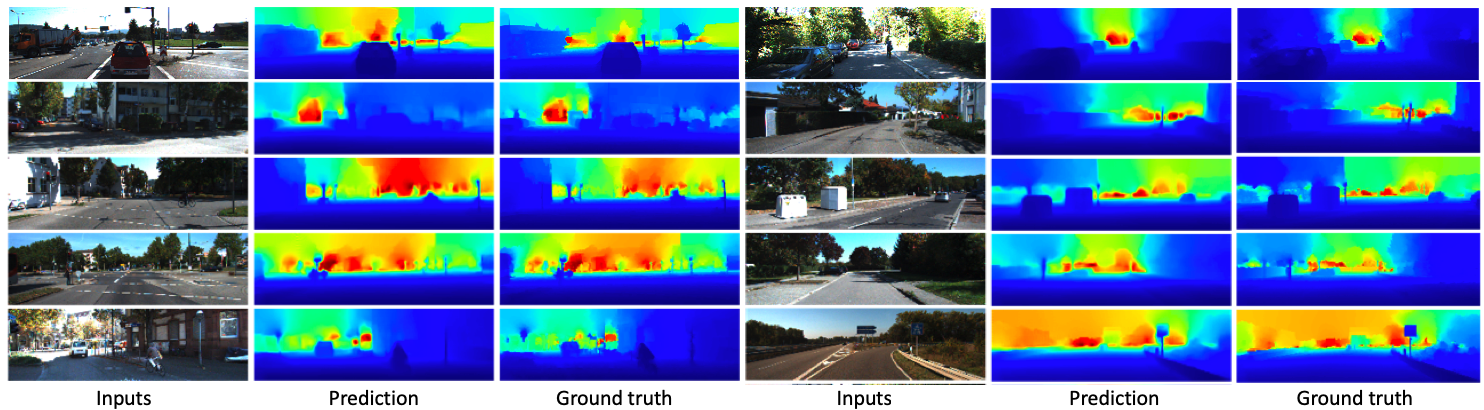}
    \end{center}
       \vspace{-12pt}
       \caption{Results on KITTI validation set.}
    \label{fig:kitti}
    \vspace{-8pt}
    \end{figure*}
    
    \begin{table*}[t]
    \centering
    \vspace{-5pt}
    \caption{Performance on KITTI validation set. All scores are evaluated on Eigen split (\cite{Eigen2015Predicting}).}
    \label{tab:kitti}
    \small
    \setlength\tabcolsep{4pt}
    \begin{tabular}{|c|c|c|c|c|c|c|c|}
    \hline
    \multirow{2}{*}{Method} & \multicolumn{4}{c|}{Error (lower is better)} & \multicolumn{3}{c|}{Accuracy (higher is better)}\\
    & Abs Rel  & Sq Rel  & RMSE  & $\text{RMSE}_{\log}$   & $\sigma < 1.25$ & $\sigma < 1.25^2$ & $\sigma < 1.25^3$ \\ \hline
    \cite{Saxena2009Make3D}   & 0.280     & 3.012  & 8.734   & 0.361     & 0.601  & 0.820   & 0.926 \\ \hline
    \cite{Liu2016Learning}   & 0.217     & 1.841 & 6.986   & 0.289     & 0.647  & 0.882    & 0.961 \\ \hline 
    \cite{Zhou2017Unsupervised} & 0.208     & 1.768      & 6.858   & -  & 0.678   & 0.885 & 0.957    \\ \hline
    \cite{Eigen2014Depth}    & 0.190     & 1.515      & 7.156   & 0.270     & 0.692   & 0.899  & 0.967   \\ \hline
    \cite{Garg2016Unsupervised} & 0.177 &1.169 & 5.285 & - & 0.727& 0.896 & 0.962 \\ \hline
    \cite{Kundu2018AdaDepth}   & 0.167     & 1.257      & 5.578   & 0.237    & 0.771   & 0.922    & 0.971   \\ \hline
    \cite{Zhan2018Unsupervised} &0.135 &1.132& 5.585& 0.229& 0.820& 0.933& 0.971 \\ \hline
    \cite{Godard2017Unsupervised}   & 0.114     & 0.898      & 4.935   & 0.206     & 0.861   & 0.949   & 0.976  \\ \hline
    \cite{Kuznietsov2017Semi}   & 0.113     & 0.741      & 4.621   & 0.189     & 0.862  & 0.960  & 0.986  \\ \hline
    \textbf{Ours}    &\textbf{0.097}    & \textbf{0.398}     &\textbf{3.007}  & \textbf{0.133}   &\textbf{0.913} & \textbf{0.985}  & \textbf{0.997} \\ \hline
    \end{tabular}
    \vspace{-6pt}
    \end{table*}
    
    \section{Experiments}
    \label{sec:experiments}
    In this section, we present our experiments on two large-scale datasets by introducing the implementation details, benchmark performance, and ablation studies validating the effectiveness of the proposed approach.
 
    \textbf{Implementation Details} \hspace{0.3cm} The proposed method is implemented using the TensorFlow 1.10 framework and runs on a single NVIDIA TITAN X GPU with 12 GB memory. The encoder-decoder structure from both stage 1 and stage 2 are identical but without weight sharing. The depth auto-encoder is trained from scratch, while the image encoder is initialized with ImageNet (\cite{Olga2015ImageNet}) pre-trained parameters. For multi-scale feature fusion, we consider four levels of feature maps which are derived from different blocks of the DenseNet-121 backbone with the feature map sizes 1/4, 1/8, 1/16 and 1/32 of the input images. For instance, in NYU Depth V2 dataset, with the input resolution $480 \times 640$, four feature maps with cascading sizes $120\times160$, $60\times80$, $30\times40$, $15\times20$ are extracted.
    The network is trained with initial learning rate $0.001$, and decreased every 10 epochs. The weight decay and momentum set to $10^{-6}$ and $0.9$ respectively. We used the Adam optimizer and batch normalization during training, with normalization decay 0.97. We set the weights for each objective as $\lambda_{depth}=1$, $\lambda_{gradient}=1$, $\lambda_{norm}=1$, and $\lambda_{CMRC}=2$. The gradient loss is added after 4k steps of training, and the surface normal loss is added after 8k steps of training.  
    \begin{figure*}[t]
    \begin{center}
    \includegraphics[width=\textwidth,height = 4cm]{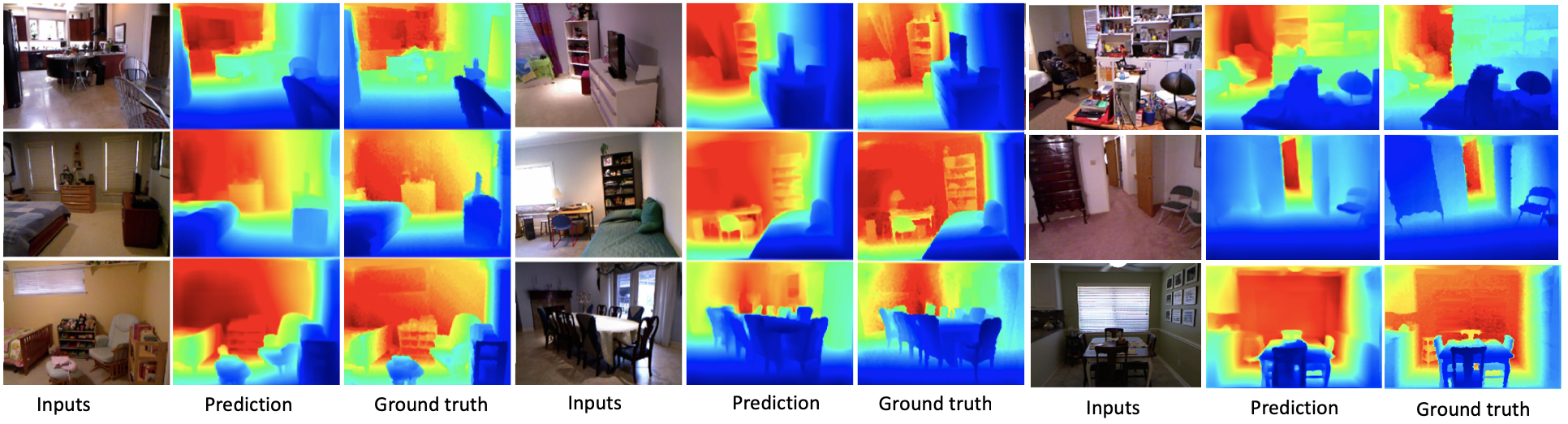}
    \end{center}
    \vspace{-15pt}
       \caption{Examples of predicted depth maps on NYU V2 Depth dataset.}
       \vspace{-8pt}
    \label{fig:NYU Depth V2}
    \end{figure*}
    
    \textbf{Data Augmentation} \hspace{0.3cm} We employ several data augmentation techniques on NYU Depth V2 dataset to prevent overfitting from limited amount of data, including: (i) \textit{Random Cropping} by $0-10\%$ of the image height/ width; (ii) \textit{Scaling} the original image by the factor interval of $[0.75, 1.25]$; (iii) 
    \textit{Random Flipping} $50\%$ of the images horizontally; (iii) \textit{Rotating} the images randomly with the degree of $[-10^\circ, 10^\circ]$; (iv) \textit{Color jitter} of brightness (by -10 to 10 of original value), contrast (by a factor of 0.5 to 2.0), saturation and hue (by -20 to 20 of original value). 
    
    \textbf{Evaluation Metrics} \hspace{0.3cm} Below is a list of evaluation metrics the quantitative evaluation is performed: 
    (1) the absolute mean relative error (Abs Rel): $\frac{1}{N}\sum_{i \in N}\frac{|d_i - d^*_i|}{d^*_i}$,
    (2) the squared relative error (Sq Rel): $\frac{1}{N}\sum_{i \in N}\frac{||d_i - d^*_i||^2}{d^*_i}$,
    (3) the root mean squared error (RMSE): $\sqrt{\frac{1}{N}\sum_{i \in N}||d_i - d^*_i||^2}$,
    (4) log mean squared error ($\text{RMSE}_{\log}$): $\sqrt{\frac{1}{N}\sum_{i\in N}||\log(d_i) - log(d_i^*)||^2}$, 
    (5) average log 10 error (Avg $\log_{10}$): $\frac{1}{N}\sum_{i \in N}|\log_{10}(d_i) - \log_{10}(d_i^*)|$, and
    (6) accuracy with threshold $t$ (t=$1.25$, $1.25^2$, $1.25^3$): $\frac{1}{N}\sum_{i\in N}1_{\{\delta = max(\frac{d_i^*}{d_i}, \frac{d_i}{d_i^*}) < t\}}$.
    
     \textbf{Results on KITTI Dataset  (Eigen split)} \hspace{0.3cm} The KITTI dataset is a large scale dataset for autonomous driving, which contains depth images captured with LiDAR sensor mounted on a driving vehicle. In our experiment, to compare the results at the same level, we follow the experimental protocol proposed by \cite{Eigen2015Predicting}, in which around $22600$ images (resolution $384 \times 1280$) from $32$ scenes are utilized as training data, and around $800$ images from $29$ scenes are used for validation. Following the previous works, the depth value of the RGB image is scaled to 0-80m. During training, the depth maps are down-scaled to resolution $192 \times 640$, and up-sampled to the original size in evaluation process. Table \ref{tab:kitti} shows the comparison with the state-of-the-art methods on KITTI dataset. We compared with state-of-the-art methods (\cite{Saxena2009Make3D,Liu2015Deep,Zhou2017Unsupervised,Eigen2014Depth,Garg2016Unsupervised,Kundu2018AdaDepth,Zhan2018Unsupervised,Godard2017Unsupervised,Kuznietsov2017Semi}). Particularly, the  methods proposed by \cite{Saxena2009Make3D,Liu2015Deep,Zhou2017Unsupervised,Eigen2014Depth,Kundu2018AdaDepth} only employ monocular images in both training and testing, while approaches in  \cite{Zhan2018Unsupervised,Garg2016Unsupervised,Kuznietsov2017Semi,Godard2017Unsupervised} are unsupervised methods that use stereo images in training and apply single image during testing. The proposed method outperforms all these methods by a large margin, and Figure \ref{fig:kitti} displays a few visualized prediction results on examples randomly chosen from the validation dataset.\\
    \indent\textbf{Results on NYU Depth V2 Dataset} \hspace{0.3cm} The NYU Depth V2 dataset contains 120K pairs of RGB-D (resolution $480 \times 640$) captured by Kinect. The dataset is manually selected and annotated into 1449 RGB-D pairs, in which 795 images are used for training, and the rest for validation. The depth value ranges from 0 to 10m. In the training process, the depth maps are down-scaled to resolution $120 \times 160$, and in testing/ evaluation, the predicted depth map is upsampled to the original resolution. Table \ref{tab:nyu} shows the comparison of the proposed method with state-of-the-art methods (official test split). We compare with both hand-crafted feature based approaches (\cite{Saxena2009Make3D,Kevin2012Depth,Shi2014Pulling}) and deep learning based ones (\cite{Liu2014Discrete,Zhuo2015Indoor,Li2015Depth,Wang2015Towards,Xu2018PAD,Liu2016Learning,Roy2016Monocular}). Figure \ref{fig:NYU Depth V2} shows examples of predicted depth maps on the NYU Depth V2 dataset.
    
    \begin{table}[t]
    	\vspace{-8pt}
    	\caption{Performance on NYU Depth V2. $\delta_1: \sigma<1.25, \delta_2: \sigma<1.25^2,\delta_3: \sigma<1.25^3$.}
    	\label{tab:nyu}
    	\centering
    	\resizebox{0.49\textwidth}{!}
    	{
    		\begin{tabular}{|c|c|c|c|c|c|c|}
    			\hline
    			\multirow{2}{*}{Method}                         & \multicolumn{3}{c|}{Error} & \multicolumn{3}{c|}{Accuracy}        \\ \cline{2-7} 
    			& Rel        & RMSE         & $\log_{10}$       & $\delta_1$ & $\delta_2$ & $\delta_3$ \\ \hline
    			\cite{Saxena2009Make3D} & 0.349   & 1.214        & -     & 0.447     & 0.745     & 0.897   \\ \hline
    			\cite{Kevin2012Depth}  & 0.35           & 1.2     & 0.131 & - & -      & -   \\ \hline
    			\cite{Liu2014Discrete}  & 0.335  & 1.06 & 0.127 & - & - & - \\ \hline
    			\cite{Shi2014Pulling}   & -  & -  & -  & 0.542  & 0.829  & 0.941             \\ \hline
    			\cite{Zhuo2015Indoor}  & 0.305  & 1.04  & -  & 0.525  & 0.838  & 0.962             \\ \hline
    			\cite{Li2015Depth} & 0.232  & 0.821  & 0.094  & 0.621  & 0.886  & 0.968   \\ \hline
    			\cite{Wang2015Towards}  & 0.220  & 0.745  & -  & 0.605  & 0.890  & 0.970  \\ \hline
    			\cite{Xu2018PAD} & 0.214 &0.792& 0.091& 0.643& 0.902 &0.977 \\ \hline
    			\cite{Liu2016Learning}  & 0.213  & 0.759  & 0.087 & 0.650 & 0.906 & 0.976 \\ \hline
    			\cite{Roy2016Monocular}  & 0.187  & 0.744  & - & - & - & - \\ \hline
    			\textbf{Ours} ($E_i + D_{pure}$) & \textbf{0.231}  & \textbf{0.828} & \textbf{0.095} & \textbf{0.631} & \textbf{0.889} & \textbf{0.968} \\ \hline
    			\textbf{Ours} ($E_i + D_{FPN}$) & \textbf{0.229}  & \textbf{0.803} & \textbf{0.092} & \textbf{0.633} & \textbf{0.891} & \textbf{0.969} \\ \hline
    			\textbf{Ours} ($E_i + D_{FPN} + align$) & \textbf{0.148}  & \textbf{0.627} & \textbf{0.075} & \textbf{0.802} & \textbf{0.944} & \textbf{0.986} \\ \hline
    			\textbf{Ours} ($E_i + D_{FPN} + SOM$) & \textbf{0.136}  & \textbf{0.604} & \textbf{0.067} & \textbf{0.814} & \textbf{0.959} & \textbf{0.990} \\ \hline
    	\end{tabular}}
    	\vspace{-12pt}
    \end{table}
    \textbf{Ablation Studies} \hspace{0.3cm} To further demonstrate the effectiveness of the proposed method, we conduct ablation studies from two aspects on NYU Depth V2 dataset. Firstly, we compare the performance of the depth estimation pipeline with different decoder structures: (1) The decoder that simply uses symmetric structure with the encoder that cascadingly upsample the feature map until the output size. (2) The decoder that takes four different feature maps from the encoder and fuses them in a pyramid fashion (as described in Section \ref{subsec:ae}). The qualitative comparison are shown in Table \ref{tab:nyu} ($E_i+D_{pure}$ and $E_i+D_{FPN}$). As can be seen from the evaluation results, the decoder structure with pyramid multi-sacle feature fusion out-performs the one that only takes the latent feature as input by a large margin, especially in the $\delta
    _1 < 1.25$ metric. Therefore, it is obvious that the mixture of features from different levels are beneficial for the details compensation (i.e. contour, edges).
  
    To validate the effectiveness of the proposed SOM module, we compare the performance of the proposed method with SOM settings against direct alignment and analyze the results. Firstly, we add the feature alignment loss for latent feature maps based on the $E_i+D_{FPN}$ structure to test the performance of direct feature alignment ($E_i + D_{FPN} + align$). The quantitative results of direct alignment rarely improved compared with the one that is trained without feature alignment loss, reflecting the limited capability of the encoder for feature adaptation. Then, we add the SOM module at feature level ($E_i + D_{FPN} + SOM$) and compare the results with the baseline structure that goes without memory. The large margin quantitative improvement in Table \ref{tab:nyu} implies that structure-specific feature alignment with memory mechanism (SOM) is superior to other approaches such as direct alignment. 
    
    \section{Conclusion}
    In this paper, we developed a novel memory guided network named Structure-Attentioned Memory Network for monocular depth estimation, consisting of the encoder-decoder based structure, as well as the external SOM module which is trained to learn and memorize the structure attentioned image-depth-residual pattern in cross-modality latent alignment. The proposed method achieves state-of-the-art performance on challenging large-scale benchmarks, and each component is validated to be effective in the ablation study. 
    
    \bibliographystyle{aaai}
    \bibliography{egbib}
    
    \end{document}